\documentclass[10pt,twocolumn,letterpaper]{article}
\usepackage{iccv}
\usepackage{my}

\makeatletter
\@namedef{ver@everyshi.sty}{}
\makeatother

\iccvfinalcopy 

\begin{document}

\title{{\SSTplus}: Effective Multi-Source Pretraining for 3D Indoor Scene Understanding}

\author{
{Yu-Qi Yang}$^{1}$\thanks{Interns at Microsoft Research Asia. \textsuperscript{\dag}Contact person.}  \qquad
{Yu-Xiao Guo}$^{2}$\qquad
{Yang Liu}$^{2\dag}$ \\
\small {$^1${Tsinghua University}\quad \quad
  $^2${Microsoft Research Asia}} \\
{\tt\small \{t-yuqyan, yuxgu, yangliu\}@microsoft.com}
}

\maketitle

\begin{abstract}
Data diversity and abundance are essential for improving the performance and generalization of models in natural language processing and 2D vision. However, 3D vision domain suffers from the lack of 3D data, and simply combining multiple 3D datasets for pretraining a 3D backbone does not yield significant improvement, due to the domain discrepancies among different 3D datasets that impede effective feature learning. In this work, we identify the main sources of the domain discrepancies between 3D indoor scene datasets, and propose {\SSTplus}, an enhanced architecture based on {\SST} for efficient pretraining on multi-source 3D point clouds. {\SSTplus} introduces domain-specific mechanisms to {\SST}'s modules  to address domain discrepancies and enhance the network capability on multi-source pretraining.
Moreover, we devise a simple \textit{source-augmentation} strategy to increase the pretraining data scale and facilitate supervised pretraining. We validate the effectiveness of our design, and demonstrate that {\SSTplus} surpasses the state-of-the-art 3D pretraining methods on typical indoor scene understanding tasks. Our code and models will be released at \url{https://github.com/microsoft/Swin3D}.
\end{abstract}



\section{Introduction} \label{sec:intro}

Point clouds are a common representation of 3D data, and extracting point-wise features from them is crucial for many 3D understanding tasks. Deep learning methods have achieved impressive results and progress in this domain, but they often require large and diverse datasets to improve feature learning. This is a common strategy in natural language processing and 2D vision, and it has also proven to be effective in enhancing model performance and generalizability in 3D vision~\cite{Wang2021,Xie2020,Zhang2021a,yang2023swin3d}. However, 3D data is much more scarce and less annotated than image and text data, which poses significant challenges and limitations for the development and impact of 3D pretraining.

A straightforward solution to mitigate the data scarcity problem is to merge multiple existing 3D datasets and use the merged data to pretrain a universal 3D backbone. However, this solution overlooks the domain differences among different 3D point clouds, such as point densities, point signals, and noise characteristics. These differences can adversely affect pre-training quality and even reduce performance, as the work of PPT~\cite{wu2023towards} observed. To address this issue, we conduct a comprehensive analysis of the domain discrepancy among 3D indoor scene datasets and identify key factors that may affect multi-source pretraining: varied data sparsity and varied signal variation across different datasets.

Based on our analysis on domain discrepancy, we present {\SSTplus}, a novel architecture that extends the {\SST} framework for multi-source pretraining, which tackles the domain discrepancy problem. Our main contributions are:
(1) we design domain-specific mechanisms for {\SST}, such as \emph{domain-specific voxel prompts} that handle the sparse and uneven voxel distribution across domains and enhance the sparse-voxel feature attention in a multi-source setting, the \emph{domain-modulated contextual relative signal embedding} scheme that captures the domain-specific signal variations and improves the contextual relative signal embedding using a tensor-decomposition-based representation, \emph{domain-specific initial feature embedding} and \emph{domain-specific layer normalization} that capture the data-source
priors in a separate way; (2) we use a source-augmentation strategy that leverages different kinds of signals from the datasets to flexibly increase the amount of training data and boost the network pretraining.

We conducted supervised multi-source pretraining of {\SSTplus} on two indoor scene datasets with different data characteristics: Structured3D~\cite{zheng2020structured3d} and ScanNet~\cite{dai2017scannet}. We tested the performance and generalizability of {\SSTplus} on various downstream tasks, such as 3D semantic segmentation, 3D detection, and instance segmentation. The results demonstrate that {\SSTplus} outperforms the state-of-the-art methods and achieves:  78.2 mIoU on ScanNet segmentation (Val), 64.1 mAP @0.5 on ScanNet 3D detection, 80.2 mIoU on 6-fold S3DIS segmentation, and 60.7 mAP @0.5 on S3DIS 3D objection detection. We also performed comprehensive ablation studies to validate the effectiveness of our architectural design. Moreover, we demonstrated that finetuning the domain-specific parameters of {\SSTplus} is a powerful and efficient strategy for data-efficient learning, achieving a significant improvement over existing approaches.

\section{Related Work} \label{sec:related}

\subsection{3D Point Cloud Pretraining}
In this section, we provide a brief overview of the recent developments in point cloud pretraining, focusing on three key aspects: \textit{backbone architectures}, \textit{pretraining schemes}, and \textit{datasets}. We also review the existing literature on multi-source training.

\paragraph{Backbone architectures}
Sparse-voxel-convolution~\cite{Wang2017,choy20194d,Graham2018} are efficient in memory and computation, and have been widely used as the core component in many 3D pretraining works built upon U-Net~\cite{Long2015} or HRNet~\cite{hrnet} architectures. However, recent works have demonstrated that transformer architectures adapted to point clouds~\cite{guo2021pct,zhao2021point,wu2022point,park2022fast} or sparse voxels~\cite{lai2022stratified} can enhance feature learning through attention on large receptive fields, and have been applied to 3D pretraining~\cite{Yu_2022_CVPR,Pang2022MaskedAF,yang2023swin3d}. Our work is based on the architecture of {\SST}~\cite{yang2023swin3d}, which employs an efficient self-attention mechanism on sparse voxels that reduces the memory complexity from quadratic to linear, and we improve this approach for effective multi-source pretraining.

\paragraph{Pretraining schemes}
Many existing 3D pretraining methods employ self-supervised learning strategies, which do not require labeled data. The early approaches~\cite{Yang2018,Wu2016} use Autoencoder or Generative Adversarial Networks to learn latent representations of 3D data, via reconstructing data or data distribution. Later methods~\cite{Xie2020,Wang2021,hou2020efficient,Zhang2021a} adopt contrastive learning, which encourage the similarity between positive pairs and the dissimilarity between negative pairs of 3D data. Transformer models, which are powered by masked signal modeling, such as BEiT~\cite{bao2022beit} and masked autoencoder~\cite{HeCXLDG22}, have also been extended to 3D pretraining~\cite{Liu2022maskdis,zhang2022point,Yu_2022_CVPR,Pang2022MaskedAF,wu2023masked,yan20233d}. Recently, supervised 3D pretraining has benefited from the rapid advancement of synthetic data generation, for instance, {\SST}~\cite{yang2023swin3d}, is pretrained on a large synthetic dataset~\cite{zheng2020structured3d} in a supervised manner, outperforms the previous pretraining methods by a large margin.

\paragraph{Datasets} Most existing methods rely on ShapeNet~\cite{shapenet2015} and ScanNet~\cite{dai2017scannet} for pretraining, but these datasets have limited diversity and size: ShapeNet contains only \SI{57}{k} CAD models from 55 common man-made categories, while ScanNet offers semantic segmentation annotations for about \SI{1.5}{k} indoor scenes. Other 3D datasets, such as ABC~\cite{Koch2019} (one million CAD models) and 3RScan~\cite{Wald2019} (1482 RGBD scans), are seldom used for pretraining. Swin3D~\cite{yang2023swin3d} demonstrate that large-scale data is essential for the performance of the pretrained backbone and use Structured3D~\cite{zheng2020structured3d}, a synthetic indoor scene dataset that is 10 times larger than ScanNet. Previous studies also indicate that (1) the backbone pretrained on CAD models suffers from a domain gap when applied to scene-level downstream tasks, as reported by PointContrast~\cite{Xie2020}, and (2) the performance improvement from large pretraining data depends on the capability of backbone architecture, as shown by Swin3D~\cite{yang2023swin3d}.

\paragraph{Multi-source training}
Multi-source training data can enhance model peformance, but it also poses several challenges. One challenge in supervised learning is to handle the inconsistent label taxonomies across different datasets. Previous works in computer vision have addressed this challenge by various methods, such as aligning data taxonomies and correcting incompatible annotations~\cite{Lambert2020}, unifying label spaces with pseudo labeling~\cite{Zhao2020}, or using dataset-specific training protocols, losses, and outputs while sharing a common architecture~\cite{Zhou2021}. The work of PPT~\cite{wu2023towards} proposes a \emph{language-guided categorical alignment} method to align the label spaces of multiple datasets for 3D point cloud pretraining, which we adopt in our multi-source supervised pretraining. Another challenge is to deal with the domain discrepancy, where multiple datasets have different data characteristics. A common solution is to apply \emph{domain-specific normalization} to the network architecture, such as using separate batch normalization and instance normalization layers for each domain~\cite{seo2019learning,chang2019domain,huang2023normalization}. PPT~\cite{wu2023towards} applied domain-specific batch normalization to their convolution-based backbone for multi-source 3D pretraining.  Our work introduces domain-specific 3D self-attention and other modules that effectively reduce the domain discrepancy.


\section{Domain Discrepancy Analysis} \label{sec:gap}

Different factors, such as scene locations, capture devices, reconstruction algorithms, and annotation methods, influence the data characteristics of 3D indoor scene datasets. These characteristics include point density and noise pattern, color quality, scene types and diversity, and annotation level and precision. These variations result in a \textit{domain discrepancy} (or \textit{domain gap}) among different datasets. \cref{tab:gap} compares some data characteristics of a few representative 3D indoor scene datasets.
In this section, we propose to use the \emph{$\mH$-divergence}~\cite{ben2010theory} tool to quantify the domain discrepancy among different datasets \cref{subsec:divergence}, and choose two important factors:  window sparsity and signal variation to further analyze the source of domain discrepancy in \cref{subsec:sparsity,subsec:signal}. These analyses guide our network design for multi-source pretraining.

\begin{table*}[t]
	\centering
	\begin{tabular}{@{}cccll@{}}
	\toprule
	\multicolumn{1}{c}{\textbf{Dataset}} &
	\multicolumn{1}{c}{\textbf{Data type}} &
	  \multicolumn{1}{c}{\textbf{Capture device}} &
	  \multicolumn{1}{c}{\textbf{Point signal}} &
	  \multicolumn{1}{c}{\textbf{Scene type}} \\ \midrule
	ScanNet~\cite{dai2017scannet}      & real data  & Structure sensor          & P+C+N & commercial and residential rooms        \\
	S3DIS~\cite{S3DIS}        & real data & Matterport camera         & P+C+N & buildings of educational and office use \\
	Structured3D~\cite{zheng2020structured3d} & synthetic data  & virtual scan              & P+C+N & professional house design
	\\ \bottomrule
	\end{tabular}
	\caption{Domain characteristics of point clouds of common 3D indoor scene datasets. The table excludes the differences in data annotation. P, C, and N stand for point coordinate, color, and normal, respectively. } \label{tab:gap}
\end{table*}

\begin{table}[t]
	\centering
		\begin{tabular}{@{}cc@{}}
			\toprule
			\mythead{Datasets}      & \mythead{$d_{\mH}(\mS, \mT)$}                                 \\ \midrule
			ScanNet  \& Structured3D                &  1.994                         \\
	
			ScanNet \& S3DIS                     &  1.976                         \\
			\bottomrule
		\end{tabular}
		\caption{$\mH$-divergence metrics of two different datasets, including \{Structured3D, ScanNet\} and \{ScanNet, S3DIS\}.  }\label{tab:divergence}
\end{table}

\subsection{Evaluation of domain discrepancy} \label{subsec:divergence}

\emph{$\mH$-divergence} is a measure of the difference between two datasets $\mS$ and $\mT$. It is based on the idea that if there is a large divergence between $\mS$ and $\mT$, then there should exist a classifier $h: \mx \rightarrow \{0, 1\}$ that can accurately separate the samples from $\mS$ and $\mT$. The classifier $h$ belongs to a set of domain classifiers $\mH$, where each classifier assigns the label $0$ to the sample $\mx_\mS$ from $\mS$, and $1$ to the sample $\mx_\mT$ from $\mT$. The formal definition of $\mH$-divergence is:
\begin{equation}
    d_{\mH}(\mS, \mT) = 2 \bigl( 1 - \min_{h \in \mH} \left( \mathrm{err}_\mS(h(\mx)) + \mathrm{err}_\mT(h(\mx))\right)\bigr),
\end{equation}
where $\mathrm{err}_\mS$ and $\mathrm{err}_\mT$ are the prediction errors of the classifier $h$ for a given $\mx$ sampled from $\mS$ and $\mT$, respectively. The $\mH$-divergence quantifies the domain discrepancy that can be captured by domain classifiers. It has a range from $[0,2]$ (negative values are unlikely to happen), with $0$ indicating that no classifier can distinguish between the two datasets, and $2$ indicating that there exists a perfect classifier that can separate them completely. A higher value of $\mH$-divergence implies a larger degree of domain discrepancy.

We conducted an experiment to measure the $\mH$-divergence of two 3D indoor scene datasets: Structured3D, a large synthetic dataset, and ScanNet, a real scanning dataset. We used two domain classifiers trained on the original training split of both datasets: one with a 3D Sparse CNN encoder structure~\cite{choy20194d} and another with a \SST-S encoder architecture~\cite{yang2023swin3d}. As the task is relatively simple, in both network architectures, we use a simple multi-resolution encoder to extract the global feature of each scene. In each resolution, we only use two Sparse CNN Resblocks or two {\SST} blocks followed by a downsample layer. For each training batch, we ensured a balanced data distribution from the two datasets ($1:1$) and randomly cropped a \SI{5}{m^3} cubic region from the input point cloud as the encoder input. To estimate the $\mH$-divergence, we used a fixed set of cropped point cloud regions from the validation sets of both datasets and computed the prediction errors of the classifiers. As shown in \cref{tab:divergence}, there is a significant domain discrepancy between Structured3D and ScanNet. We also measured the $\mH$-divergence between ScanNet and S3DIS~\cite{S3DIS}, which are both real-world datasets, and found that it is slightly lower than that between Structured3D and ScanNet. This indicates that the domain discrepancy between real-world datasets is also significant.

\subsection{Window sparsity}  \label{subsec:sparsity}
Many 3D deep learning models for 3D point cloud learning adopt either 3D convolution or self-attention mechanisms to process local regions formed by sparse voxels.  The number of occupied voxels in each region depends on both the point sampling method and the 3D content of the scene. To investigate how this number varies across different 3D indoor scene datasets, we conduct the following analysis.

\begin{figure*}[t]
	\centering
	\begin{overpic}[width=0.49\columnwidth]{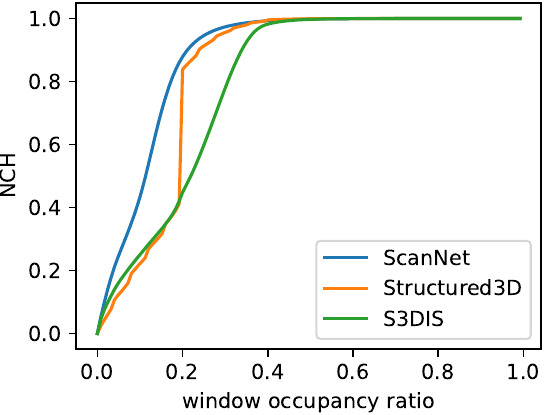}
	\put(10,0){\small\textbf(a)}
	  \end{overpic}
	  \hfill
	\begin{overpic}[width=0.49\columnwidth]{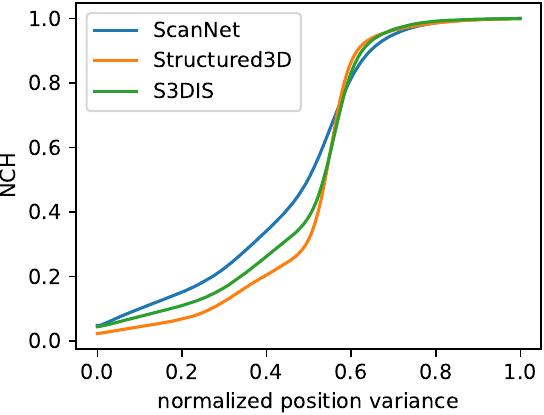}
	\put(10,0){\small\textbf(b)}
	  \end{overpic} \hfill
		\begin{overpic}[width=0.49\columnwidth]{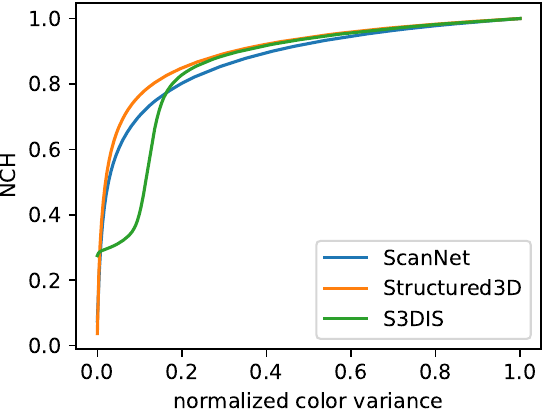}
	\put(10,0){\small\textbf(c)}
	  \end{overpic} \hfill
		\begin{overpic}[width=0.49\columnwidth]{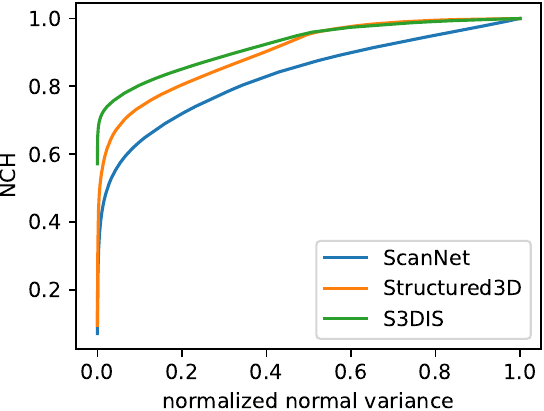}
	\put(10,0){\small\textbf(d)}
	  \end{overpic}
	\caption{Window sparsity and signal variation analysis on windows of size $5\times 5 \times 5$. The voxel size is set to \SI{2}{cm}. (a) The  normalized cumulative histogram (NCH) of the ratio of occupied voxels in the window. (b), (c), and (d) represent the NCHs of the variances of the positions, colors, and normals of the points in the window, respectively. All statistical calculations are based on the average of 200 scenes for each dataset.  The variance ranges are normalized to $[0, 1]$ in the figures.}
	\label{fig:sparsity}
  \end{figure*}

We voxelize the point cloud of a 3D scene into a sparse grid with a uniform voxel size of \SI{2}{cm}. We then partition the grid into non-overlapping windows of size $5\times 5 \times 5$. For each non-empty window $w$, we calculate its \emph{window occupancy ratio} $w_{or}$, which is the fraction of occupied voxels in $w$. The normalized cumulative histogram (NCH) of $w_{or}$ reveals the sparsity pattern of the scene at the window level. \cref{fig:sparsity}-(a) shows the averaged NCH for three datasets: ScanNet, Structure3D, and S3DIS. We can see that these datasets have different degrees of window sparsity. In particular, the NCH curve of Structured3D has a sharp increase at $w_{or}=\frac{1}{5}$. This is because Structured3D contains many clean vertical and horizontal planes that fill up $5\times 5$ voxels within a $5\times 5 \times 5$ window.

\subsection{Signal variation} \label{subsec:signal}

The input to the 3D neural networks often consists of the relative changes of point signals, such as position, color, and normal. To examine how these relative changes vary across different datasets, we propose the following evaluation strategy.

We adopt the same configuration as described in \cref{subsec:sparsity}, and compute the variances of point position, color, and normals within a window $w$, denoted by $\Var_p$, $\Var_c$, and $\Var_n$ respectively: $\Var_{p}  = \frac{1}{2N^2} \sum_{x, y\in w} \|\bm{p}(x) - \bm{p}(y)\|^2;
    \Var_{c} = \frac{1}{2N^2}\sum_{x, y\in w} \|\bm{c}(x) - \bm{c}(y)\|^2;
    \Var_{n}  = \frac{1}{2N^2}\sum_{x, y\in w} \|\bm{n}(x) - \bm{n}(y)\|^2$.
Here, $\bm{p}$, $\bm{c}$, and $\bm{n}$ represent point coordinate, color, and normal respectively, while $N$ is the number of non-empty voxels in $w$. Following existing 3D learning approaches, we resample the input point cloud to ensure that each non-empty voxel contains only one point. To assess the distribution of signal variance, we use the normalized cumulative histogram (NCH). \cref{fig:sparsity}-(b,c,d) show the averaged NCH for three datasets (ScanNet, Structure3D, and S3DIS) with respect to $\Var_p$, $\Var_c$, and $\Var_n$ respectively. The position variance NCH curve of Structured3D indicates that it has more diverse geometric changes in shape than the other two datasets. Moreover, the color variance NCH curve of Structured3D reveals that its synthetic color richness is still lower than the other two real datasets. Lastly, the lower NCH curve of normal variance for ScanNet suggests that its normal data is \emph{noisier} than the other datasets. \looseness=-1

\section{Methodology} \label{sec:method}

In this section, we first provide a concise overview of the {\SST} network architecture\cite{yang2023swin3d} (\cref{subsec:swin3d}) and then introduce our novel modifications to {\SST} that enable effective multi-source pretraining (\cref{subsec:swin3dplus}).

\begin{figure*}[t]
	\centering
	\begin{overpic}[width=0.85\textwidth]{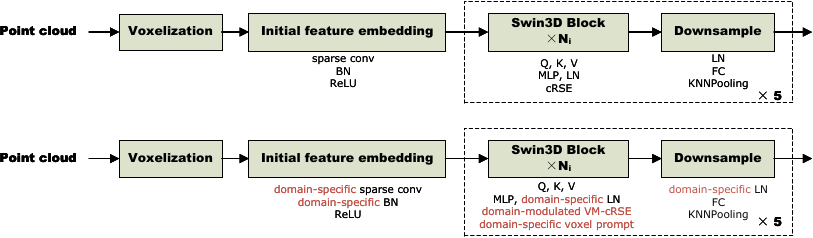}
	  \put(0,16){\small \textbf{(a)} {\SST}}
	  \put(0,2){\small \textbf{(b)} {\SSTplus} }
	  \end{overpic}
	\caption{Overview of network architectures of  {\SST} and {\SSTplus}.  $N_1, N_2, N_3, N_4, N_5 = 2, 4, 9, 4, 4$.  BN, LN and FC refer to \textit{batch normalization}, \textit{layer normalization} and \textit{fully-connected layer}. Q, K, V are Key, Query and Value tensors of self-attention.
	}
	\label{fig:swinarch}
  \end{figure*}

\subsection{Recap of {\SST}} \label{subsec:swin3d}
{\SST} is a novel 3D backbone architecture for point cloud analysis that adopts the Swin Transformer\cite{liu2021swin} model. It uses a hierarchical network structure that converts the input point cloud into a multi-scale sparse voxel grid.
The initial feature embedding layer transforms the raw voxel features into a high-dimensional space using sparse convolution. Then, the voxel features are processed by a sequence of {\SST} blocks and down-sampled at different scales. Each {\SST} block applies memory-efficient self-attention to the sparse voxel features within regular or shifted windows. To enhance the feature learning capability, its self-attention mechanism employs \emph{contextual relative signal encoding} (cRSE), which utilizes learnable look-up tables to map the relative signal changes to high-dimensional features and integrates these content-aware features into the self-attention computation.  \cref{fig:swinarch}-(a) shows the {\SST} network architecture. We briefly review the key components of  {\SST} below.

\paragraph{Input} The input point cloud $\mS$ consists of points with feature vectors $\ms_{\mp} \in \mathbb{R}^M$, which include point coordinates and other signal information such as point color and normal if available.

\paragraph{Voxelization} $\mS$ is voxelized into a sparse voxel grid with five levels of resolution, where the finest voxel size is \SI{2}{cm} for indoor scenes. For each nonempty voxel, a \textit{representative point} is chosen from the points within the voxel, and the voxel feature is set as the feature of its \textit{representative point}.

\paragraph{Initial feature embedding} A sparse convolution layer with kernel size 3 is applied to the voxels at the finest level, followed by a BatchNorm (BN) layer and a ReLU layer. The voxel feature that is used for the convolution is the concatenation of the positional offset to the voxel center and other point signals.

\paragraph{{\SST} block}  The {\SST} block performs self-attention on regular and shifted windows in 3D, with window size $M \times M \times M$. It adopts the standard transformer architecture, consisting of multi-head self-attention, LayerNorm layers (LN) and an MLP layer. It enhances its self-attention with a memory-efficient implementation and a contextual relative signal encoding scheme (cRSE), and the computation formula is given as follows (we omit the multi-head notation for simplicity).

Let $\{\mf_i\}_{i=1}^N$ denote the feature vectors of nonempty voxels $\{\mv_i\}_{i=1}^N$ in a given window. Linear projection tensors $\bQ, \bK, \bV$ are defined for \emph{Query}, \emph{Key}, and \emph{Value}, respectively.
The self-attention mechanism computes the output feature vector $\mf^\star_{i}$ at each voxel $\mv_i$ as follows:
\begin{equation}
    \mf^\star_{i} = \dfrac{\sum_{j=1}^N \exp(e_{ij}) (\mf_j \bV +  \mt_{V}\bigl(\Delta \ms_{ij})\bigr)}{\sum_{j=1}^N \exp(e_{ij})},  \label{eq:2}
\end{equation}
where $e_{ij}$ is the scaled attention score between the query vector $\mf_i \bQ$ and the key vector $\mf_j \bK$, given by:
\begin{equation}
    e_{ij} = \dfrac{(\mf_i \bQ) (\mf_j \bK)^T + b_{ij}}{\sqrt{d}}, \label{eq:eij}
\end{equation}
with $d$ being the channel dimension. The term $b_{ij}$ is the contextual relative signal encoding of the displacement vector $\Delta \ms_{ij} := \ms_{\mv_i}-\ms_{\mv_j}$, which captures the contextual relation between the query and the key voxels. It is computed as:
\begin{align}
    b_{ij} = (\mf_i \bQ) \bigl(\mt_{K}(\Delta \ms_{ij})\bigr)^T +  (\mf_j \bK) \bigl(\mt_{Q}(\Delta \ms_{ij})\bigr)^T.   \label{eq:3}
\end{align}
Here, $\mt_{V}, \mt_{K}, \mt_{Q}$ are trainable functions that map the bounded $\Delta \ms_{ij}$ to
$\mathbb{R}^d$.  These functions are implemented as look-up tables, following the approach of \cite{lai2022stratified,wu2021rethinking}.

\paragraph{Downsample} The Downsample module of {\SST} performs two steps. First, it applies a LayerNorm layer and a fully connected (FC) layer to each voxel feature vector. Second, it uses KNNpooling to aggregate the features of the nearest neighbors for the downsampled voxels. \looseness=-1

\paragraph{Limitations of naive multi-source pretraining} Pretraining {\SST} on multiple datasets by combining data alone does not guarantee better performance on downstream tasks. This is evident from the results of {\SST} pretrained on Structured3D only and {\SST} pretrained on both ScanNet and Structured3D. The former outperforms the latter on the downstream task of ScanNet semantic segmentation, as shown in \cref{tab:bad}. This finding is consistent with \cite{wu2023towards}, which also observed lower performance of SparseUNet pretrained on three datasets: ScanNet, Structured3D and S3DIS, without considering domain discrepancy.

\begin{table}[t]
    \centering
    \begin{tabular}{@{}cccc@{}}
        \toprule
        \mythead{Net} & \mythead{single-source} & \mythead{multi-source} \\ \midrule
        \SST         &    \textbf{76.8}             &      76.5                \\
        SparseUNet    &     \textbf{75.8}           &      73.6                \\
        \bottomrule
    \end{tabular}
    \caption{Naive multi-source pretraining does not lead better performance on ScanNet segmentation (validation set). The results of SparseUNet are from Tab.~2(a) \& Tab.~3 of \cite{wu2023towards}. The reported values are mean IoUs of segmentation.} \label{tab:bad}
\end{table}

\subsection{Design of {\SSTplus}} \label{subsec:swin3dplus}
To address the aforementioned issues in multi-source pretraining, we introduce {\SSTplus}, a novel approach that enhances the {\SST} model with several domain-specific components. These components are designed to isolate the domain-specific features from the shared backbone, resulting in better performance and generalization across different domains. The components include: \emph{domain-specific initial feature embedding} (\cref{subsubsec:virtual}),  \emph{domain-specific layer normalization} (\cref{subsubsec:norm}),
\emph{domain-specific voxel prompts} (\cref{subsubsec:virtual}), \emph{domain-modulated cRSE} (\cref{subsubsec:crse}) and its enhanced version --- \emph{domain-modulated VM-cRSE} (\cref{subsubsec:vmcrse}),
and a \emph{source augmentation} strategy (\cref{subsubsec:aug}).  We illustrate the main differences between our approach and the original {\SST} architecture in \cref{fig:swinarch}-(b).

We use the notation $\{\mD_l\}_{l=1}^L$ to denote the multiple datasets that we use for pretraining, where $L$ is the number of datasets. We allow different input signal types for different datasets. For instance, the point clouds in $\mD_1$ may only have point coordinates, while the point clouds in $\mD_2$ may have both point coordinates and color information.

\subsubsection{Domain-specific initial feature embedding} \label{subsubsec:embedding}
We use a separate feature embedding module for each data source to obtain the initial embeddings of the raw point features. This allows the module to capture the data-source priors and properties during the pretraining process. We refer to this module as \emph{domain-specific}, meaning that the network contains $L$ different feature embedding modules during the pretraining. This design slightly increases the network parameters, as the feature embedding module is lightweight.

\subsubsection{Domain-specific layer normalization} \label{subsubsec:norm}
To take domain discrepancy into account, we introduce domain-specific layer normalization (DSLN) for {\SST}, inspired by the concept of domain-specific batch normalization and instance normalization~\cite{seo2019learning}. DSLN allows each data source to have its own element-wise affine transformation in the LN layer, instead of sharing the same parameters. The DSLN formula for data $\mS \in \mD_l$ is given by:
\begin{equation}
    \mf^\star = \frac{\mf-\mathrm{E}[\mf]}{\sqrt{\mathrm{Var}[\mf]+\epsilon}} \cdot \gamma_l + \beta_l,
\end{equation}
where $\mathrm{E}[\mf]$ and $\mathrm{Var}[\mf]$ are the mean and variance of all voxel features, computed over channel dimension. The parameters of $\gamma_l$ and $\beta_l$ are domain-specific parameters, and $\epsilon$ is a small constant to prevent numerical instability. This modification is simple but effective, and has not been explored by previous work for both multi-source training and pretraining.

\subsubsection{Domain-specific voxel prompts} \label{subsubsec:virtual}
\begin{figure}[t]
	\centering
	\begin{overpic}[width=0.8\columnwidth]{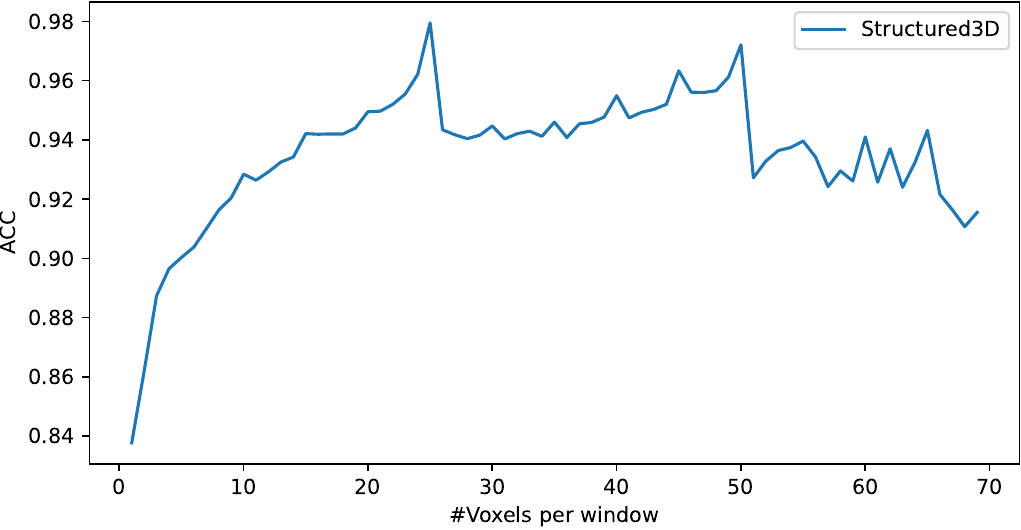}
	\end{overpic}
	\caption{Segmentation accuracy of {\SST} on the Structure3D validation set. The x-axis represents the count of nonempty voxels in a $5\times5 \times 5$ grid with a voxel size of \SI{2}{cm}. The y-axis indicates the average segmentation accuracy across windows with an equal number of nonempty voxels. }
	\label{fig:acc}
\end{figure}

Since {\SST} calculates self-attention between nonempty voxels within a window, we hypothesize that windows with very few voxels have limited information exchange, and the learned voxel features in such windows may be less effective. To validate this hypothesis, we evaluate the segmentation accuracy of {\SST} on the Structured3D dataset. We compute the average voxel accuracy for windows with the same number of nonempty voxels and visualize the accuracy curve in \cref{fig:acc}. The findings clearly demonstrate that windows containing less than $10$ nonempty voxels exhibit considerably lower accuracy in comparison to others.

We address the above issue across different datasets by introducing a set of virtual voxels $\{\tilde{\mv}_{b}\}_{b=1}^B$ into each {\SST} block, which are referred to as \emph{domain-specific voxel prompts}. These virtual voxels are not shared by different {\SST} blocks, and each virtual voxel $\tilde{\mv}_{b}$ is associated with a learnable and domain-specific feature vector $\{\tilde{\mf}_{b}\}$. This helps to reduce the extreme window sparsity issue and provides data-source prior to guide the feature learning. The self-attention formula is revised to take into account these additional voxels and features:
\begin{equation}
    \begin{split}
        \mf^\star_{i} = & \biggl(\sum_{j=1}^{N} \exp(e_{ij}) (\mf_j \bV +  \mt_{V}\bigl(\Delta \ms_{ij})\bigr) +            \\
                        &  \sum_{b=1}^{B} \exp(\tilde{e}_{ib}) (\tilde{\mf}_b \bV \bigr) \biggr) /                          \\
                        & \bigl(\sum_{j=1}^{N} \exp(e_{ij}) + \sum_{b=1}^{B} \exp(\tilde{e}_{ib})\bigr), \quad i = 1, \ldots, N,
    \end{split} \label{eq:6}
\end{equation}
where $e_{ij}$ is the same to \cref{eq:eij} and
\begin{equation}
    \tilde{e}_{ib} = \dfrac{(\mf_i \bQ) (\tilde{\mf}_b  \bK)^T }{\sqrt{d}}, \quad b = 1, \ldots, B  \label{eq:7}
\end{equation}
Here note that there is no contextual relative signal encoding between virtual voxels and real voxels in the above formulae as the virtual voxels are not bind with any raw features.
In our implementation, we empirically choose $B=5$ to strike a balance between performance and efficiency.

\subsubsection{Domain-Modulated cRSE} \label{subsubsec:crse}
To encode the raw signal differences that vary across different datasets, as demonstrated in \cref{subsec:signal}, the cRSE in {\SST} needs to be adapted for each dataset during pretraining. A straightforward solution is to use a \emph{domain-specific} cRSE, but this would increase the parameter size of {\SST} rapidly, making it difficult to fit the network of {\SST} in GPUs with limited memory when pretraining with multiple datasets. Therefore, we propose an alternative solution, called \emph{domain-modulated cRSE}, which leverages a shared base cRSE for multiple datasets and applies a domain-specific modulation to the base cRSE.

\paragraph{Base cRSE} The base cRSE is identical to the original cRSE, and we use the same symbols $\mt_{V}, \mt_{K}, \mt_{Q}$ to represent them. We express the signal difference $\Delta \ms$ between voxel $\bm{u}$ and $\bm{v}$ as a vector: $(\delta_1, \ldots, \delta_m)$, corresponding to the three components of coordinate, color, and normal, respectively.
We write the base cRSE as follows:
\begin{equation}
    \mt_R(\Delta \ms) = \sum_{m=1}^M \mt_m(\delta^q_m), \quad R \in \{Q,K,V\}  \label{eq:8}
\end{equation}
where $\delta^q_m$ is a quantized version of $\delta_m$ and $\mt_m$ is a trainable look-up table that maps $\delta^q_m$ to the feature space.

\paragraph{Domain modulation} In each {\SST} block, we introduce a set of domain-specific look-up tables $\{\mt_{m}^l, l = 1, \ldots, L\}$ to modulate the base cRSE.
\begin{equation}
    \mt_{R,l}(\Delta \ms) = \sum_{m=1}^M \mt_{m}^l(\delta^q_m) \cdot \mt_m(\delta^q_m), \quad R \in \{Q,K,V\}
    \label{eq:9}
\end{equation}
where $\mt_{m}^l$ maps $\delta^q_m$ to a scalar value.
To apply the domain-aware cRSE formulation, we only need to substitute $\mt_{V}, \mt_{K}, \mt_{Q}$ in \cref{eq:6} with $\mt_{V,l}, \mt_{K,l}, \mt_{Q,l}$.  The additional parameters introduced by all $\mt_{m}^l$s per {\SST} block are tiny, as they amount only to $3 \times M \times L \times T $, where $T$ is the number of quantization divisions and 3 refers to \{Q,K,V\}. When cRSE uses point coordinate, color and normal, $M$ is equal to 9.

\subsubsection{Domain-Modulated VM-cRSE}  \label{subsubsec:vmcrse}
To encode relative signal changes optimally, one should account for the correlation of the changes in all dimensions. A possible implementation is to use a look-up table with a $T^M$ grid for mapping the quantized changes. This approach is called the \emph{Product method} in the previous work\cite{wu2021rethinking}. However, this method suffers from high storage cost when $M$ is large and low utilization rate of the grid table, as the signal changes are sparse in $\mathbb{R}^M$ space. A simpler approximation of this method is the \emph{cross method} in ~\cite{wu2021rethinking}, which is given by \cref{eq:8}, but it has a limited ability to capture correlation among different signal dimensions.

\paragraph{VM-cRSE}
We take a different view at the $T^M$ grid table mentioned above: the grid can be regarded as a Tensor with dimension $T \times T \times \cdots \times T \times C$, where $C$ is the feature dimension. Without loss of generality, we assume $M=9$ which represents the changes of point coordinates, colors and normals. Inspired by Vector-Matrix (VM) decomposition \cite{Chen2022ECCV}, we propose to approximate the grid table by VM in the following way, with the balance between efficiency and capability.

We first disregard the correlation between different types of signals, for instance, the red color channel and the x-direction of the normal, as they have no deep correlation in general. Thus, the large grid table can be replaced by three different $T^3$ grid tables: $\mathcal{T}_{p}, \mathcal{T}_{c}, \mathcal{T}_{n}$.  We utilize VM decomposition to approximate them:
\begin{equation}
    \begin{split}
        \mathcal{T}_{r}(\delta^q_{r}) = &
        \mt_{r,1}(\delta^q_{r,1}) \cdot \mt_{r,23}(\delta^q_{r,2}, \delta^q_{r,3}) +                                                           \\ &\mt_{r,2}(\delta^q_{r,2}) \cdot \mt_{r,31}(\delta^q_{r,3}, \delta^q_{r,1}) +  \\
                                        & \mt_{r,3}(\delta^q_{r,3}) \cdot \mt_{r,12}(\delta^q_{r,1}, \delta^q_{r,2}), \quad r \in \{p, c, n\}.
    \end{split} \label{eq:10}
\end{equation}
Here, $\{1,2,3\}$ in the subscripts denote the first, second and third component of the corresponding signal; $\mt_{r,k}$ and $\mt_{r,kj}$ are 1D and 2D look-up tables that maps the relative 1D and 2D changes to the feature space.

With the above VM decomposition, \cref{eq:8} can be replaced by:
\begin{equation}
    \mt_R(\Delta \ms) = \sum_{r \in \{p, c, n\}}\mathcal{T}_{r}(\delta^q_{r}), \quad R \in \{Q,K,V\}.
\end{equation}
We call the above formulation --- \emph{VM-cRSE}.

\paragraph{Domain-modulated VM-cRSE} To enable multi-source pretraining, we extend the method described in \cref{subsubsec:crse} to incorporate \emph{domain-awareness} into the VM-cRSE: we use a collection of domain-specific 1D and 2D look-up tables $\{\mt_{r,k}^l, l = 1, \ldots, L\}$ and $\{\mt_{r,kj}^l, l = 1, \ldots, L\}$ to adjust the base cRSEs. The equation in \cref{eq:10} is modified as follows:
\begin{equation}
    \small
    \begin{split}
         & \mathcal{T}^l_{r}(\delta^q_{r}) =
        \mt^l_{r,1}(\delta^q_{r,1}) \cdot \mt_{r,1}(\delta^q_{r,1})  \cdot  \mt^l_{r,23}(\delta^q_{r,2}, \delta^q_{r,3}) \cdot \mt_{r,23}(\delta^q_{r,2}, \delta^q_{r,3})
        \\ & +\mt^l_{r,2}(\delta^q_{r,2}) \cdot \mt_{r,2}(\delta^q_{r,2}) \cdot  \mt^l_{r,31}(\delta^q_{r,3}, \delta^q_{r,1}) \cdot \mt_{r,31}(\delta^q_{r,3}, \delta^q_{r,1})
        \\& +
        \mt^l_{r,3}(\delta^q_{r,3}) \cdot \mt_{r,3}(\delta^q_{r,3})  \cdot \mt^l_{r,12}(\delta^q_{r,1}, \delta^q_{r,2}) \cdot \mt_{r,12}(\delta^q_{r,1}, \delta^q_{r,2}), \\   &r \in \{p, c, n\}.
    \end{split}
\end{equation}
$\mt_{r,k}^l$ and $\mt_{r,kj}^l$ map the quantized signal changes (1D or 2D) to scalar values.
Domain-modulated VM-cRSE is by default used in {\SSTplus}. During the training phase, the shared look-up tables are initialized by a Guassian distribution whose mean is 0 and variane is 0.02, and all look-up tables for domain modulation are initialized to $1$.

\subsubsection{Source augmentation} \label{subsubsec:aug}
To analyze point clouds, we often need to consider different types of signals associated with the points, such as coordinates, color, normal, and voxel occupancy. However, most existing methods use a fixed set of signals as input to the network, which limits their generalization ability to handle data with varying signal types. For example, if the input data have more or fewer signals than expected, one may have to discard or fabricate some signals, and retrain or finetune the network accordingly.

On the other hand, our multi-source pretraining setting allows us to leverage the different types of signals from multiple point clouds to enhance the pretraining performance. We introduce a source-augmentation strategy to expand the pretraining datasets, as follows. Without loss of generality, we assume that a dataset $\mA$ consists of three types of signals: point coordinates (p), color (c), and normal (n). We can create multiple datasets from $\mA$ with various combinations of signals: $\mA_p, \mA_{pc}, \mA_{pn}, \mA_{pcn}$. By mixing these augmented datasets with other datasets, we can increase the quantity and diversity of data for pretraining without extra cost.  The enlarged datasets are beneficial to pretrain the shared parameters of the networks, and signal-specific data features can be learned by our domain-specific network design.

\paragraph{Adaption for domain-modulated cRSE} In order to facilitate domain-modulated cRSE for augmented sources with different signal types, we introduce virtual point signal types that may not be present in certain augmented datasets. These virtual point signal types are used solely for cRSE computation, and we assume that the added point signal values are identical. This ensures that the relative signal changes are always zero, allowing the domain-modulated look-up tables to only map the zero change to a learnable scalar value.


\section{Experimental Analysis} \label{sec:results}

In this section, we first present the detailed configuration of the multi-source pretraining of {\SSTplus} (\cref{subsec:pretraining}), and demonstrate the effectiveness of the proposed modules through comprehensive ablation studies (\cref{subsec:ablation}). In \cref{subsec:downstream}, we report the performance of the finetuned {\SSTplus} on typical downstream tasks on indoor scenes, and demonstrate its advantage over existing methods in both fully-annotated and weakly-annotated settings.


\subsection{Multi-source pretraining} \label{subsec:pretraining}

\paragraph{Network configuration} Swin3D~\cite{yang2023swin3d} proposed two versions of pretrained models: {\SST-S} and {\SST-L}. Both versions have the same configuration of window sizes and block layers: the first stage has a window size of $5 \times 5 \times 5$ and the rest have $7 \times 7 \times 7$, and the layer numbers are $\{2,4, 9, 4, 4\}$. However, {\SST-L} has more feature channels and attention heads than {\SST-S}, which improves its performance but also increases its data and resource requirements. We designed {\SSTplus} based on the structure of {\SST-S}, aiming to balance performance and pretraining cost on multiple datasets. We set the quantization division numbers to $16$ for 1D look-up tables and $4 \times 4$ for 2D look-up tables, respectively.  \cref{tab:parameter} compares the parameter sizes of {\SST-S}, {\SST-L} and {\SSTplus}, where the shared parameters of {\SST} blocks of {\SSTplus} is larger than {\SST}-S due to the use of 2D tables.

\begin{table*}[t]
    \centering
    \begin{tabular}{@{}cccc|c@{}}
        \toprule
        \mythead{Net} & \mythead{FE.}(M) & \mythead{blocks}(M) & \mythead{Other}(M) & \mythead{Total}(M)\\
        \midrule
        \SST-S      &   0.01          &   22.87      &  0.69    &  23.57           \\
        \SST-L      &   0.01         &   58.82        &  0.69    &  60.75           \\
        \SSTplus    &   $L\times0.01$      & $L\times1.79+46.22$    & $L\times 0.01+0.69$ & $L\times 1.81+46.91$   \\
        \bottomrule
    \end{tabular}
    \caption{Net parameters of individual modules. $L$ represents the number of datasets.
    \textbf{FE}, \textbf{blocks}, \textbf{Other} refer to \emph{initial feature embedding}, \emph{{\SST} blocks}, and \emph{all the remaining modules}, respectively. } \label{tab:parameter}
\end{table*}

\paragraph{Pretraining setting}
We use Structured3D~\cite{zheng2020structured3d} (\SI{21}{K} synthetic rooms) and ScanNet~\cite{dai2017scannet} (\SI{1.5}{K} scanned rooms) as our pretraining datasets, and choose 3D semantic segmentation as the pretext task. We generate point clouds from Structured3D following \cite{yang2023swin3d}, with color, normal, and semantic label information for each point. However, the semantic labels of Structured3D and ScanNet are not compatible, so we cannot use them directly for supervised learning. To solve this problem, we apply \emph{language-guided categorical alignment}~\cite{wu2023towards}, which leverages the relationship between label texts to unify the label space. We use the same decoder architecture as in the pretraining of {\SST} \cite{yang2023swin3d} to predict the semantic labels. The only modification is that we replace the LayerNorm in decoder with a \emph{domain-specific} one to handle multi-source inputs.

\paragraph{Training details}  We follow the original data split of Structured3D and ScanNet. To balance the two datasets, we use a ratio of $2:1$, meaning that for every three batches, two contain only Structured3D data and one contains only ScanNet data. We augment the input data by randomly cropping and rotating them.
 We train the network for 100 epochs on 8 NVIDIA V100 GPUs (\SI{32}{GB}), using the AdamW optimizer with a Cosine learning rate scheduler, and taking 6 days.

\paragraph{Network efficiency} The network inference stage is efficient in terms of GPU memory and time. It requires only {\SI{6}{GB}} of GPU memory on average, and it completes the inference in {\SI{900}{ms}} for each scene, on average.


\subsection{Ablation study} \label{subsec:ablation}
We conduct an ablation study to examine the effectiveness of the proposed modules of {\SSTplus} for multi-source pretraining. We compare different pretraining settings of {\SSTplus} as follows:
\begin{enumerate} [leftmargin=*] \setlength\itemsep{1mm}
    \item [(1)] Baseline. We use the {\SST-S} network structure and pretrain it only on Structured3D data.
    \item [(2)] Baseline + Multi-source. We pretrain {\SST-S} on both Structured3D and ScanNet with \emph{language-guided categorical alignment}~\cite{wu2023towards}.
    \item [(3)] Based on (2), we introduce \emph{domain-specific initial feature embedding}.
    \item [(4)] Based on (3), we apply \emph{domain-specific layer normalization}.
    \item [(5)] Based on (4), we replace the shared cRSE with \emph{domain-modulated cRSE}.
    \item [(6)] Based on (5), we replace \emph{domain-modulated cRSE} with \emph{domain-modulated VM-cRSE}.
    \item [(7)] Based on (6), we add \emph{domain-specific voxel prompt}.
    \item [(8)] Based on (7), we enable \emph{source augmentation}. This is our default setting of {\SSTplus}.
\end{enumerate}

We test our pretraining setting on the 3D semantic segmentation task, using ScanNet (validation set) and S3DIS (Area 5) as the benchmarks. We measure the performance by the mean IoU metric. It is important to note that the S3DIS dataset was not used for the pretraining, so the performance on this dataset reflects the generalization ability of the pretrained backbones.
We follow the setup of {\SST} on the 3D segmentation task, which adds a {\SST} block at each level of the decoder after the skip-connection. We also activate the domain-specific modules in the decoder if they are activated in the corresponding {\SST++} encoder in the ablation study. The input signals of points in the benchmark include point coordinate, normal and color.
For finetuning, we load the shared parameters and the domain-specific module parameters corresponding to the ScanNet dataset if they exist. We finetune the models for 600 epochs on ScanNet and 3000 epochs on S3DIS.

\begin{table}[t]
	\centering
	\resizebox{\linewidth}{!}{
	  \scriptsize
	  \begin{tabular}{@{}ccccccc|cc@{}}
		\toprule
		\multicolumn{7}{c}{\mythead{Pretraining settings}} & \multicolumn{2}{c}{\mythead{Segmentation mIoU}(\%)}                                                                                                                    \\ \midrule
		\mythead{ID}                                       & \mythead{MS}                                        & \mythead{DI} & \mythead{DL} & \mythead{DM-cRSE} & \mythead{VP} & \mythead{SA} & \mythead{ScanNet} & \mythead{S3DIS} \\  \midrule

		(1)                                                & \xmark                                              & \xmark       & \xmark       & \xmark           & \xmark       & \xmark       & 76.8              & 72.5            \\
		(2)                                                & \checkmark                                          & \xmark       & \xmark       & \xmark           & \xmark       & \xmark       & 77.0              & 72.8            \\
		(3)                                                & \checkmark                                          & \checkmark   & \xmark       & \xmark           & \xmark       & \xmark       & 77.0              & 73.0            \\
		(4)                                                & \checkmark                                          & \checkmark   & \checkmark   & \xmark           & \xmark       & \xmark       & 77.2              & 73.2            \\

		(5)                                                & \checkmark                                          & \checkmark   & \checkmark   & DM       & \xmark       & \xmark       & 77.3              & 73.2            \\

		(6)                                                & \checkmark                                          & \checkmark   & \checkmark   & VM       & \xmark       & \xmark       & 77.6              & 73.4            \\
		(7)                                                & \checkmark                                          & \checkmark   & \checkmark   & VM        & \checkmark   & \xmark       & 77.8              & 74.0            \\
		\rowcolor{gray!20} (8)                             & \checkmark                                          & \checkmark   & \checkmark   & VM        & \checkmark   & \checkmark   & \textbf{78.2}     & \textbf{74.9}   \\
		(9)                                                & \checkmark                                          & \checkmark   & \xmark       & VM        & \checkmark   & \checkmark   & 77.6              & 73.9            \\
		\bottomrule
	  \end{tabular}
	}
	\caption{Ablation study of {\SSTplus} modules. \textbf{MS}, \textbf{DI}, \textbf{DL}, \textbf{DM-cRSE}, \textbf{VP}, \textbf{SA} refer to using multi-source datasets, domain-specific initial feature embedding, domain-specific LayerNorm, domain-modulated cRSE (DM) or domain-modulated VM-cRSE (VM), domain-specific voxel prompt and source augmentation. } \label{table:ablation}
\end{table}

The results reported in \cref{table:ablation}  show that the performance increases gradually as we adopt more components from our default setting. In particular, we found that: \emph{domain-modulated VM-cRSE} contributes significantly to the performance on ScanNet; \emph{source augmentation} and \emph{domain-specific voxel prompt} lead to the most noticeable improvement on S3DIS. We also observe that \emph{domain-specific layer normalization} plays a crucial role in boosting the performance of other modules: when we use a shared LayerNorm across different sources, the results deteriorate significantly, as shown in \cref{table:ablation}-(9), compared to (8).  We notice that the improvement achieved by \emph{domain-modulated cRSE} is marginal (see (4)\&(5)), because it has few trainable parameters. However, its improved variant \emph{domain-modulated VM-cRSE} is more effective (see (4)\&(6)), as it has more parameters that are derived from 2D look-up tables.

\begin{figure}[t]
	\centering
	\begin{overpic}[width=\columnwidth]{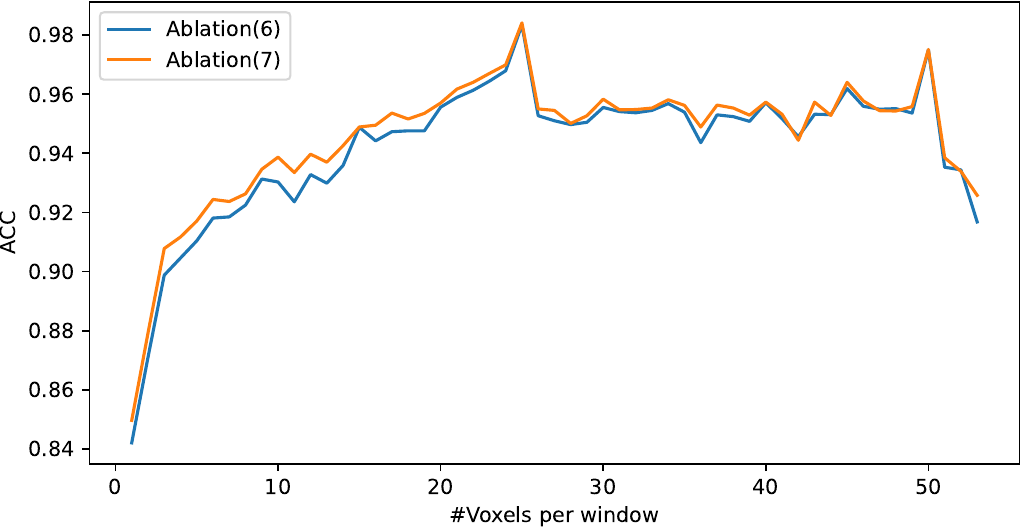}
	\end{overpic}
	\caption{Effect of \emph{domain-specific voxel prompt} on the segmentation accuracy  on the Structure3D validation set.
	}
	\label{fig:ablation_voxel_prompt}
\end{figure}

To assess the impact of the \emph{domain-specific voxel prompt} on prediction accuracy, especially on windows with less nonempty voxels, we conduct a similar evaluation as in \cref{fig:acc} on the pretrained {\SSTplus} model with the ablation settings (6) and (7). \Cref{fig:ablation_voxel_prompt} shows that the accuracy on windows with fewer nonempty voxels is improved clearly by introducing the domain-specific voxel prompt, while the improvement on grids with more nonempty voxels is less noticeable, as there are enough nonempty voxels for feature interaction.

\subsection{Evaluation on downstream tasks} \label{subsec:downstream}

We present the results of applying the multi-source pretrained {\SSTplus} to three common indoor scene understanding tasks: 3D semantic segmentation, 3D detection, and instance segmentation. We also test the data efficiency of our method. Similar to our albation study, for all experiments, we use the same {\SSTplus} backbone that is pretrained on multiple sources and its domain-specific module parameters corresponding to the ScanNet dataset. We provide the details of our training setup and the performance of our approach in the following paragraphs.

\begin{table*}[t]
	\centering
	\resizebox{0.95\linewidth}{!}{
	  \scriptsize
	  \begin{tabular}{@{}c|c|cc|cc@{}}
		\toprule
		\multirow{2}{*}{\mythead{Method}}               & \multirow{2}{*}{\mythead{Pre.}} & \multicolumn{2}{c|}{\mythead{ScanNet Segmentation}} & \multicolumn{2}{c}{\mythead{S3DIS Segmentation}}                                                           \\
														&                                 & \mythead{Val mIoU}  (\%)                            & \mythead{Test mIoU}  (\%)                        & \mythead{Area5 mIoU}  (\%) & \mythead{6-fold mIoU} (\%) \\
		\midrule
		LargeKernel3D~\cite{chen2022scaling}            & \xmark                          & 73.5                                                & 73.9                                             & -                          & -                          \\
		Mix3D~\cite{nekrasov2021mix3d}                  & \xmark                          & 73.6                                                & 78.1                                             & -                          & -                          \\
		Stratified Transformer~\cite{lai2022stratified} & \xmark                          & 74.3                                                & 74.7                                             & 72.0                       & -                          \\
		PointTransformerV2~\cite{wu2022point}           & \xmark                          & 75.4                                                & 75.2                                             & 71.6                       & -                          \\
		OctFormer~\cite{octformer}                      & \xmark                          & 75.7                                                & 76.6                                             &                            & -                          \\
		PointVector-XL~\cite{deng2022point}             & \xmark                          & -                                                   & -                                                & 72.3                       & 78.4                       \\
		WindowNorm~\cite{wang2022window}                & \xmark                          & -                                                   & -                                                & 72.2                       & 77.6                       \\
		OneFormer3D~\cite{kolodiazhnyi2023oneformer3d}  & \xmark                          & 76.6                                                & -                                                & -                          & -                          \\
		\midrule
		DepthContrast~\cite{zhang2021self}              & ScanNet                         & 71.2                                                & -                                                & 70.6                       & -                          \\
		SceneContext~\cite{hou2020efficient}            & ScanNet                         & 73.8                                                & -                                                & 72.2                       & -                          \\
		PointContrast~\cite{Xie2020}                    & ScanNet                         & 74.1                                                & -                                                & 70.9                       & -                          \\
		MaskContrast~\cite{wu2023masked}                & ScanNet                         & 75.5                                                & -                                                & -                          & -                          \\
		{\SST}-S~\cite{yang2023swin3d}                  & Structured3D                    & 76.8                                                & -                                                & 73.0                       & 78.2                       \\
		{\SST}-L~\cite{yang2023swin3d}                  & Structured3D                    & 77.5                                                & 77.9                                             & 74.5                       & 79.8                       \\
		\midrule
		OneFormer3D~\cite{kolodiazhnyi2023oneformer3d}  & ScanNet+Structured3D            & -                                                   & -                                                & 72.4                       & 75.0                       \\
		PPT~\cite{wu2023towards}                        & ScanNet+Structured3D+S3DIS      & 76.4                                                & 76.6                                             & 72.7                       & 78.1                       \\
		PonderV2~\cite{zhu2023ponderv2}                 & ScanNet+Structured3D+S3DIS      & 77.0                                                & \textbf{78.5}                                    & 73.2                       & 79.1                       \\
		\rowcolor{gray!20} {\SSTplus}                   & ScanNet+Structured3D            & \textbf{78.2}                                       & 77.7                                             & \textbf{74.9}              & \textbf{80.2}              \\
		\bottomrule
	  \end{tabular}
	}
	\vspace{2pt}
	\caption{Quantitative evaluation on semantic segmentation.
	  \textbf{Pre.} indicates whether the method uses pretraining and which datasets are used. DepthConstrast, SceneContext, PointContrast and MaskContrast are pretrained in a self-supervised manner, and other pretraining-based methods utilize data labels. The hyphen symbol (-) denotes that the method does not perform a certain task. We report the best scores achieved by different methods.
	}  \label{tab:seg-all-in-one} \vspace{-2mm}
\end{table*}

\begin{figure}[t]
	\centering
	\includegraphics[width=0.95\columnwidth]{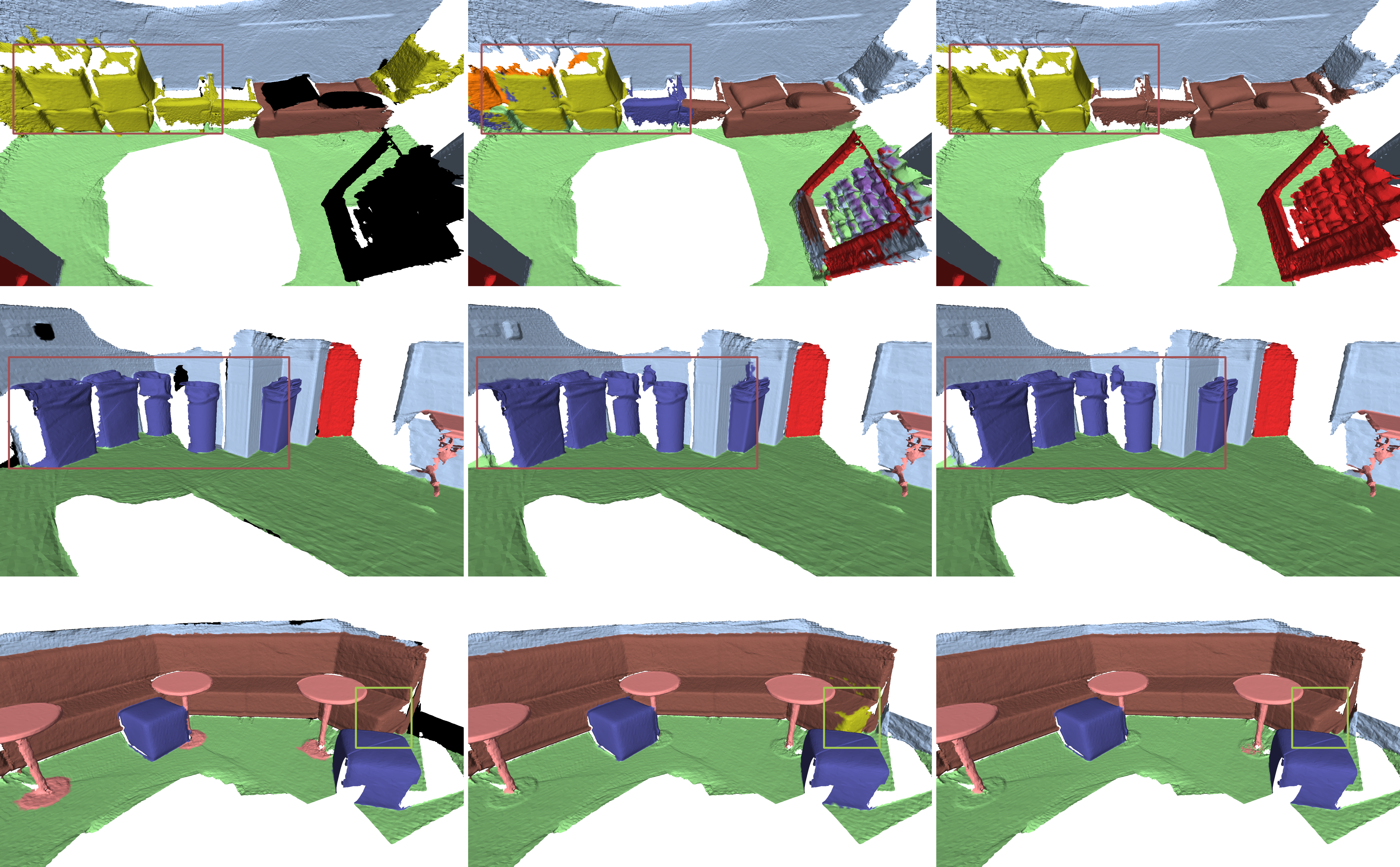}
	\caption{Visual comparison of ScanNet segmentation. \textbf{Left}: Ground-truth segmentation labels. \textbf{Middle}: {\SST-L}~\cite{yang2023swin3d}'s results. \textbf{Right}: {\SSTplus}'s results.}
	\label{fig:scannet_seg_vis}
\end{figure}

\paragraph{3D semantic segmentation}
We conduct semantic segmentation on two widely used datasets: ScanNet and S3DIS. The data splits of ScanNet are 1201 (training), 312 (validation), 100 (test); and S3DIS contains 272 scenes across 6 areas.  We adopt the same decoder and finetuning configurations as in our ablation study. We show the results on the ScanNet validation set, the ScanNet test set, Area 5 of S3DIS, and the S3DIS 6-fold cross-validation in \cref{tab:seg-all-in-one}.
We also benchmark our method against existing approaches that use different training settings:  training from scratch, pretrained with a single dataset, pretrained with multiple datasets.  As shown in \cref{tab:seg-all-in-one}, the recent pretraining-based approaches, such as {\SST-S}, {\SST-L}, Point Prompt Training (PPT)~\cite{wu2023towards}, and PonderV2~\cite{zhu2023ponderv2} which used PPT for pretraining with a reconstruction-based pretext task, outperform most unprtrained methods on both tasks. Among them, our {\SSTplus} achieves the best performance on ScanNet(val) and S3DIS, but behind {\SST-L}, Mix3D and PonderV2 on ScanNet(test). In \cref{fig:scannet_seg_vis}, we provide visual comparison of {\SST-L} and {\SSTplus} on three ScanNet(val) scenes, and highlight the improvement of {\SSTplus} over {\SST-L}.

\begin{table}[t]
    \centering
    \scriptsize
        \begin{tabular}{@{}l|c|cc@{}}
            \toprule
            \mythead{Method}                        & \mythead{Pre.} & \mythead{mAP@0.25} & \mythead{mAP@0.5} \\
            \midrule
            CAGroup3D~\cite{wang2022cagroupd}       & \xmark         & 75.1               & 61.3              \\
            + {\SST}-L~\cite{yang2023swin3d}                       & \checkmark     & 76.4               & 63.2              \\
            \rowcolor{gray!20} + {\SSTplus} & \checkmark     & 76.2               & 64.1     \\ \midrule
            V-DETR~\cite{shen2023v}                 & \xmark         & \textbf{77.8}      & \textbf{66.0}     \\
            \bottomrule
        \end{tabular}
    \vspace{2pt}
    \caption{Quantitative evaluation on 3D detection (ScanNet). }  \label{tab:detect-scannet} 
\end{table}

\begin{table}[t]
    \centering
    \scriptsize
        \begin{tabular}{@{}l|c|cc@{}}
            \toprule
            \mythead{Method}                   & \mythead{Pre.} & \mythead{mAP@0.25} & \mythead{mAP@0.5} \\
            \midrule
            FCAF3D~\cite{rukhovich2021fcaf3d}  & \xmark         & 66.7               & 45.9              \\
            + {\SST}-L~\cite{yang2023swin3d}   & \checkmark     & 72.1               & 54.0              \\
            \rowcolor{gray!20} + {\SSTplus}     & \checkmark     & \textbf{75.5}      & \textbf{60.7}     \\ \midrule
            Point-GCC+TR3D~\cite{fan2023point} & \checkmark     & 75.1               & 56.7              \\
            \bottomrule
        \end{tabular}
    \vspace{2pt}
    \caption{Quantitative evaluation of 3D detection (S3DIS).
    }  \label{tab:detect-s3dis}
\end{table}

\begin{figure}[t]
	\centering
	\includegraphics[width=0.95\columnwidth]{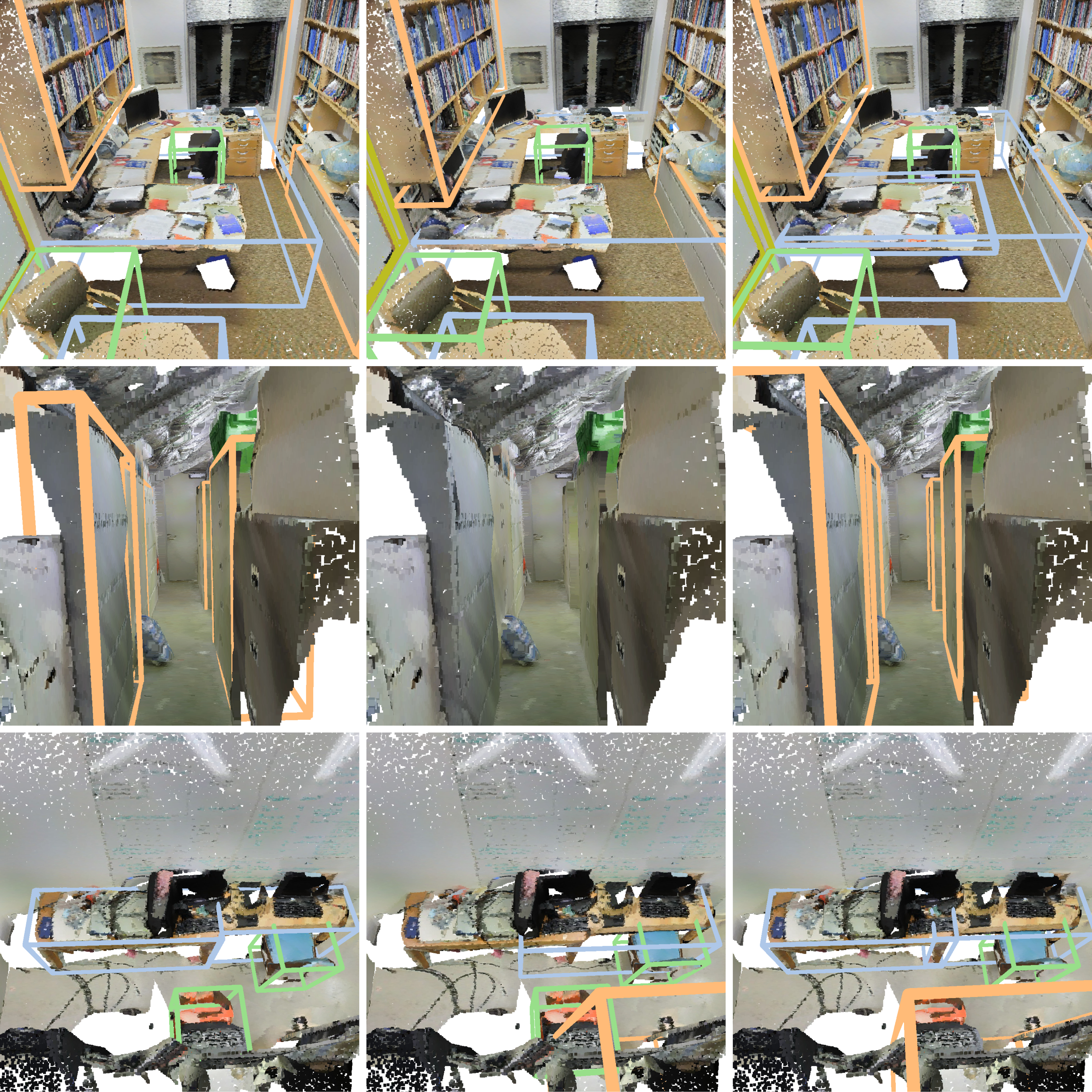}
      \vspace{2pt}
	\caption{Visual comparison of 3D Detection on S3DIS scenes. \textbf{Left}: Ground-truth object boxes. \textbf{Middle}: FCAF3D+{\SST}~\cite{yang2023swin3d}'s results. \textbf{Right}: FCAF3D+{\SSTplus}'s results.}
	\label{fig:s3dis_det_vis}
\end{figure}

\paragraph{3D detection} We adopt the same training setup and evaluation protocol as {\SST}~\cite{yang2023swin3d} to test the performance of {\SSTplus} on 3D detection. We use either CAGroup3D~\cite{wang2022cagroupd} or FCAF3D~\cite{rukhovich2021fcaf3d} as the detector architecture, and replace their original feature extractor with our pretrained {\SSTplus}. \cref{tab:detect-scannet} and \cref{tab:detect-s3dis} show the detection performance on ScanNet and S3DIS datasets, where we use mAP@0.25 and mAP@0.5 metrics to evaluate the results. We observe that the use of pretrained {\SSTplus} achieves comparable mAP@0.25 score to {\SST}, and significantly higher score on mAP@0.5. Notably, {\SSTplus}+FCAF3D achieves the state-of-the-art performance on S3DIS. \cref{fig:s3dis_det_vis} visualizes the improvement of {\SSTplus} over {\SST-L} on three S3DIS scenes. This demonstrates that multi-source pretraining via {\SSTplus} enhances 3D detection.
We also note that the detectors we employed are not the best ones, and more advanced architectures such as V-DETR~\cite{shen2023v} can achieve state-of-the-art performance. It is promising to combine our backbone with the latest detector design to further improve the performance.

\begin{table*}[t]
    \centering
    \scriptsize
    \begin{tabular}{@{}l|c|ccc|ccc@{}}
        \toprule
        \multirow{2}{*}{\mythead{Method}}     & \multirow{2}{*}{\mythead{Pre.}} & \multicolumn{3}{c|}{\mythead{ScanNet}} & \multicolumn{3}{c}{\mythead{ScanNet200}}                                                                                \\
                                              &                                 & \mythead{mAP}                          & \mythead{mAP@0.50}                       & \mythead{mAP@0.25} & \mythead{mAP} & \mythead{mAP@0.50} & \mythead{mAP@0.25} \\
        \midrule
        Mask3D~\cite{schult2022mask3d}                                & \xmark                          & \textbf{55.2}                                   & \textbf{78.0}                                     & \textbf{87.0}               &   -            &     -               &            -        \\ \midrule
        PointGroup~\cite{jiang2020pointgroup} & \xmark                          & 36.0                                   & 56.9                                     & 72.8               & 15.8          & 24.5               & 32.2               \\
        + MaskContrast~\cite{wu2023masked}    & ScanNet                         & -                                      & 59.6                                     & -                  & -             & 26.8               & -                  \\
        + {\SST}-S~\cite{yang2023swin3d}      & Structured3D                    & 39.2                                   & 61.5                                     & 76.2               & 18.9          & 29.0               & 36.1               \\
        + PPT~\cite{wu2023towards}            & ScanNet+Structured3D+S3DIS      & 40.7                                   & 62.0                                     & 76.9               & 19.4          & 29.4               & 36.8               \\

        \rowcolor{gray!20} + {\SSTplus}       & ScanNet+Structured3D            & \textbf{41.0}                          & \textbf{62.6}                            & \textbf{77.3}      & \textbf{21.1} & \textbf{32.2}      & \textbf{38.0}      \\
        \bottomrule
    \end{tabular}
    \vspace{2pt}
    \caption{Quantitative evaluation on 3D instance segmentation.
    }  \label{tab:ins-seg} \vspace{-2mm}
\end{table*}

\paragraph{3D instance segmentation}
We follow the experimental setup of MaskContrast~\cite{wu2023masked} and PPT~\cite{wu2023towards} to evaluate our approach on 3D instance segmentation, where PointGroup~\cite{jiang2020pointgroup} is the baseline method for the instance segmentation task. The original design of PointGroup uses Sparse-conv-based UNet as the backbone, which we replace with our pretrained {\SSTplus}. We also compare our backbone with the pretrained {\SST-S}. We select ScanNet and ScanNet200 datasets as testbeds, where the latter dataset is a fine-grained ScanNet dataset with 200 object categories. \cref{tab:ins-seg} shows the common metrics (mAP, mAP@0.50, mAP@0.25) of different approaches. We observe that multi-source training (PPT and {\SSTplus}) significantly improves the performance over the baseline, while {\SSTplus} is the best one among the approaches that use PointGroup architecture. However, more optimal instance segmentation designs like Mask3D outperform PointGroup, so there is an opportunity to integrate them with our pretrained backbone.

\begin{table*}[t]
    \centering
    {
        \scriptsize
        \begin{tabular}{@{}c|c|c|cccc@{}}
            \toprule
            \multirow{2}{*}{\mythead{Method}}               & \multirow{2}{*}{\mythead{Pre.}}  & \multirow{2}{*}{\mythead{FineTune Params.}} & \multicolumn{4}{c}{\mythead{ScanNet200}}  \\
            & & & \mythead{1\%} & \mythead{5\%} & \mythead{10\%}  & \mythead{20\%} \\
            \midrule
            MaskContrast~\cite{wu2023masked}                & ScanNet             & All & 4.5 & 11.7 & 16.5 & 20.3 \\
            \SST-S\cite{yang2023swin3d}                   & Structured3D        & All & 6.1 & 14.4 & 19.7 & 23.2 \\
            PPT~\cite{wu2023towards}                        & ScanNet+Structured3D+S3DIS & All & 6.9 & 14.2 & 19.0 & 22.0  \\
            \rowcolor{gray!20} {\SSTplus}                   & ScanNet+Structured3D       & All & 7.6 & 15.2 & \textbf{20.7} & \textbf{24.2}  \\
            \rowcolor{gray!20} {\SSTplus}                   & ScanNet+Structured3D       & Domain-specific & \textbf{8.0} & \textbf{16.0} & 20.4 & 23.9  \\
            \bottomrule
        \end{tabular}
    }
    \vspace{2pt}
    \caption{Data efficient learning on ScanNet200 segmentation. }  \label{tab:few_shot} \vspace{-2mm}
\end{table*}

\paragraph{Data efficient learning}

We use the ScanNet200 segmentation task~\cite{rozenberszki2022language} for data efficient learning. ScanNet200 has 200 class categories, far more than 20 categories of the original ScanNet; and the labels have not been leaked to the pretraining. We follow the data-efficient split from SceneContext~\cite{hou2020efficient} to finetune the backbone and report the mean mIoU on the validation set, in which $1\%, 5\%, 10\%, 20\%$ data are chosen as training data. Note that the $1\%$ split only has 12 scenes for training.
To avoid overfitting the limited data, we use a simple segmentation decoder like in the pretraining stage. We compare {\SSTplus} with other pretrained models, such as {\SST}~\cite{yang2023swin3d} and PPT~\cite{wu2023towards}.
The results show that {\SSTplus} achieves significant improvements over other methods, including  PPT~\cite{wu2023towards}, which is supervisedly pretrained on three datasets. Moreover, we find that finetuning only the domain-specific modules is a parameter-efficient way to adapt to the limited training data. In this setting, we only finetune the parameters related to domain-specific feature embedding, domain-specific voxel prompts, look-up tables for domain modulation, and the downsample and upsample layers, which have less than \SI{2.5}{M} parameters. This finetuning strategy further improves {\SSTplus} significantly when the training data is very scarce ($1\%$ and $5\%$).


\section{Conclusion} \label{sec:conclusion}
In this work, we addressed the challenge of integrating multiple 3D datasets with large domain discrepancies and leveraged their complementary strengths for improving 3D pretraining performance. The remarkable outcomes of {\SSTplus} attest to the benefits of multi-dataset pretraining and suggest more effective ways to utilize diverse 3D data. For future work, we intend to extend our method to multi-source outdoor datasets, which exhibit greater domain variety, complexity, and data scale.


{\small
  \bibliographystyle{ieee_fullname}
  \bibliography{reference}
}

\end{document}